\documentclass[conference, a4paper]{IEEEtran}

\makeatletter
\def\endthebibliography{%
  \def\@noitemerr{\@latex@warning{Empty `thebibliography' environment}}%
  \endlist
}
\makeatother

\IEEEoverridecommandlockouts
\usepackage{cite}
\usepackage{amsmath,amssymb,amsfonts}
\usepackage{algorithmic}
\usepackage{graphicx}
\usepackage{textcomp}
\usepackage{xcolor}
\usepackage{color}
\usepackage{comment}
\usepackage{ bbold }

\usepackage{tikz}
\newcommand*\circled[1]{\tikz[baseline=(char.base)]{
            \node[shape=circle,draw,inner sep=1pt] (char) {#1};}}

\definecolor{cadmiumgreen}{rgb}{0.0, 0.42, 0.24}

\usepackage{makecell}
\def\BibTeX{{\rm B\kern-.05em{\sc i\kern-.025em b}\kern-.08em
    T\kern-.1667em\lower.7ex\hbox{E}\kern-.125emX}}

\newcommand{\fakepar}[1]{~\\\noindent{\textit{#1.}}}
\begin{document}

\bstctlcite{IEEEexample:BSTcontrol}

\title{HURRA! Human readable router anomaly detection}

\author{
\IEEEauthorblockN{Jose M. Navarro, Dario Rossi}
\IEEEauthorblockA{Huawei Technologies Co. Ltd.}
\texttt{jose.manuel.navarro, dario.rossi\}@huawei.com}
}

\maketitle

\begin{abstract}
This paper presents HURRA, a system that aims to reduce the time spent by human operators in the process of network troubleshooting. To do so, it comprises two modules that are plugged after any anomaly detection algorithm: (i) a first \emph{attention mechanism},  that ranks the present features in terms of their relation with the anomaly and (ii) a second module able to incorporates previous \emph{expert knowledge} seamlessly, without any need of human interaction nor decisions. We show the efficacy of these simple processes on a collection of real router datasets obtained from tens of ISPs which exhibit a rich variety of anomalies and very heterogeneous set of KPIs, on which we gather manually annotated ground truth by the operator solving the troubleshooting ticket. Our experimental evaluation shows that (i) the proposed system is effective in achieving high levels of agreement with the expert, that (ii) even a simple statistical approach is able to extracting useful information from expert knowledge gained in past cases to further improve performance and finally that (iii) the main difficulty in live deployment concerns the automated selection of the anomaly detection algorithm and the tuning of its  hyper-parameters.
\end{abstract}


\section{Introduction}
Network operation and management, especially concerning troubleshooting, is still a largely manual and time-consuming process\cite{mazel2011sub}.
We recognize that keeping human operators ``in the loop'' is unavoidable nowadays, yet we argue it is also desirable.
On the one hand, the use of machine learning techniques  can help removing humans from the ``fast loop'', by  automatically processing large volumes of data to replace human tasks~\cite{dromard2016online,bhatia2019unsupervised},  and overall increasing the troubleshooting efficiency. On the other hand, data driven algorithms still benefit from human assistance (e.g., providing ground truth labels to improve the algorithm) and responsibilities (e.g., taking legal ownership of the action suggested by an algorithm), thus, it is positive to keep humans in the ``slow loop''.
Under these premises, it becomes obvious that the amount of time spent by human experts in the troubleshooting process is a  valuable resource that should be explicitly taken into account when designing anomaly detection systems~\cite{sota_human_labelless,sota_human_opprentice,sota_human_sfe,sota_human_feedback}, which is our focus in this work. As such, it is clear that the ultimate goal of anomaly detection is to produce a ``human readable'' output~\cite{sota_other_adele,sota_ad_dl1, sota_cloudDet,sota_refout,sota_beam,sota_lookout,sota_hics}
 that should, furthermore, be presented to the final users in a way that minimally alters their workflow.

\begin{figure*}[t]
\begin{center}
  \includegraphics[width=1\textwidth]{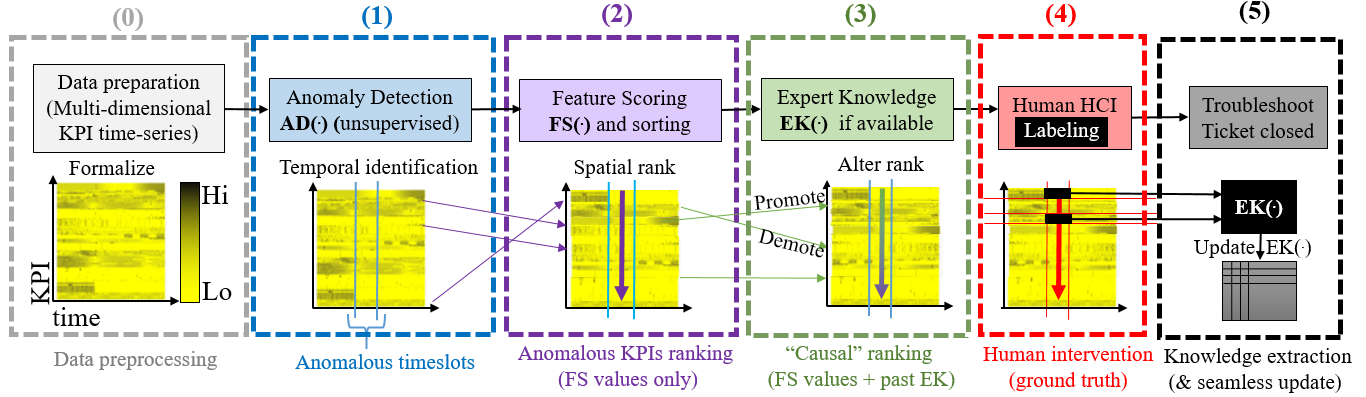}
  \caption{High-level view of the building blocks of employed  in the Human Readable Router Anomaly Detection (HURRA) system presented in this paper.}
  \label{fig:birdeye}
\end{center}
\end{figure*}

We design a system for Human Readable Router Anomaly  (HURRA) detection by combining simple building blocks, that are depicted from a very high-level perspective in  Fig.~\ref{fig:birdeye}.
From the left, router KPI data, which can be abstracted as multi-variate time series, is the input to a  \circled{0} preliminary \textit{data preprocessing} step: for instance, the heat map representation of one of our datasets shown in Fig.~\ref{fig:birdeye}, employs a standard feature regularization.
Next, \circled{1} an \textit{anomaly detection (AD)} algorithm is used to identify, along the temporal dimension, time-regions where the system  behavior presents unexpected patterns. 
Subsequently, an \circled{2} attention-focus mechanism, implemented as a  \textit{feature scoring (FS)} policy is used to rank KPIs along the spatial dimension. The purpose is to only minimally alter the existing troubleshooting system, by letting users prioritize their focus to KPIs that are likely the most relevant for troubleshooting the issue at hand.  

To further assist  human intervention, HURRA additionally leverages, if available,  \circled{3}  previous \textit{expert knowledge (EK)}  with a simple mechanism based on statistical information that alters the KPI ranking, using separate schemes to promote/demote KPIs in the ranking. By prioritizing KPIs that  experts have found to be relevant in the solution of previous troubleshooting tickets, the EK mechanism attempts at presenting KPIs that network operators more often associate with what they loosely identify as the ``root cause''  of the issue (we point out that we are not explicitly making use of causality theory\cite{pearl}, hence the quotes).  By \circled{4} explicitly flagging KPIs, experts (unconsciously) provide \textit{ground truth labels} to our system: from these labels, it is possible to \circled{5} automatically distill and \textit{update expert knowledge}, without requiring any further human interaction.

We benchmark our proposal against a private dataset comprising  manually labeled troubleshooting data, gathered from 28 operational ISPs deployments, with minute-level telemetry.  Intuitively, the main criterion to evaluate the effectiveness of HURRA is to measure the extent of agreement between the ranking of scores produced by the expert vs the algorithmic output, which is well captured by metrics such as the normalized Discounted Cumulative Gain (nDCG \cite{yahoo_rankings_from_xgboost}).

Summarizing our main contributions:
\begin{itemize}
\item We tackle the issue of attention focus on network troubleshooting, designing a decoupled AD+FS unsupervised system that can achieve a high agreement with expert labeling (nDCG up to 0.82 in our data), reducing the KPIs need to be manually verified (on average 50 KPI less in our datasets).
\item We propose a simple yet effective mechanism to account for expert knowledge, that is incremental by design and that can further improve agreement with the expert -- even when an Oracle is used to tackle the anomaly detection problem.
\end{itemize}

The rest of this paper overviews related work (Sec.~\ref{sec:related}),
describes the dataset (Sec.~\ref{sec:dataset}), details our methodology (Sec.~\ref{sec:methodology}) and experimental result (Sec.~\ref{sec:results}). We finally discuss our findings (Sec.~\ref{sec:discussion}) and summarize our work   (Sec.~\ref{sec:conclusions}).

\section{Related work}\label{sec:related}
 HURRA does not attempt to make any novel contribution for building blocks \circled{0}-\circled{1} of Fig.~\ref{fig:birdeye}, for which we employ state-of-the art techniques. Rather, leveraging unique \circled{4} human labeled datasets, we instead work on an \circled{2} attention focus mechanism that can seamlessly integrate \circled{3} expert knowledge.

\subsection{Attention focus}

Admittedly, we are not the first attempting
at ranking features~\cite{sota_other_adele,sota_ad_dl1, sota_cloudDet}, or explaining outliers~\cite{sota_refout,sota_beam,sota_lookout,sota_hics}. 

\subsubsection{Feature ranking}
For instance, \cite{sota_other_adele} ranks features based on how much they deviate from a normal distribution and extracts ``anomaly signatures'': 
while our work shares the same  spirit of~\cite{sota_other_adele}, it does not make any assumption about the underlying data distribution.
LSTM and autoencoders are exploited  in \cite{sota_ad_dl1}, where reconstruction errors
are interpreted as anomalies: while this scheme is interoperable with our,
the temporal extent and heterogenity of the data at our disposal (cfr. Sec.~\ref{sec:dataset}) do not allow us to properly train a deep neural network.

CloudDet~\cite{sota_cloudDet} instead detects anomalies through scoring  seasonally decomposed components of the time series: this couples \circled{1} and \circled{2} by assigning and scoring anomalies to each feature independently, which limits the analysis to a set of univariate features, and is thus orthogonal to our goal.

\subsubsection{Outlier explanation}
The interest for subspace methods in network domain was first raised in  \cite{mazel2011sub}, which however mostly focuses on anomaly detection, instead of explanation\cite{sota_refout,sota_beam,sota_lookout,sota_hics}. In a nutshell, these techniques crop datasets to specific feature subspaces that better represent the detected outliers, and can either tackle single (\emph{point explanation}  \cite{sota_refout,sota_beam}) or multiple  outliers (\emph{summarization explanation}~\cite{sota_lookout,sota_hics}).
RefOut~\cite{sota_refout} performs point explanation by creating random subspaces, assessing feature importance by comparing the distribution of the scores produced by an AD algorithm, increasing the number of relevant KPIs until a target dimensionality is reached. Likewise, Beam~\cite{sota_beam} builds the most interesting subspace through the maximization of the AD scores following a greedy algorithm.
Interesting contributions to summarization explanation are represented by work such as LookOut~\cite{sota_lookout} and HiCS~\cite{sota_hics}, which are similar and devise more complex statistics to characterize  interaction among features.
These algorithms could, in principle, be used to perform steps \circled{1}-\circled{2},  however they are not without  downsides. First,  outlier explanation methods assume that the points are independent from each other, which is not true for time series.
A more significant bottleneck is represented by their computational cost, which becomes prohibitive in high dimensional datasets, unlike the simple sorting in HURRA.

\subsection{Expert knowledge}
Leveraging human expertise in anomaly detection is the focus of \emph{active learning} studies such as  \cite{sota_human_labelless,sota_human_opprentice,sota_human_sfe,sota_human_feedback}, which share similarities with our work.
Authors in~\cite{sota_human_labelless} alleviate the task of generating labels for AD by using a high recall algorithm (producing many false alarm) paired with a human analysis phase, where the experts decide which anomalous ``shapes'' they are interested in finding. In our work we employ the temporal portion of the ground truth information as an AD ``Oracle'' only as a reference performance point to contrast results of AD algorithms.
An Operator's Apprentice is proposed in~\cite{sota_human_opprentice} to study univariate time series: humans are required to label a collection of datasets to train an apprentice (Random Forest), on a set of features extracted from the time series, iterating until a configurable performance threshold is attained. Similarly, \cite{sota_human_sfe} introduces an iterative process (referred to as Sequential Feature Explanation), with the explicit objective of quantifying the number of features needed to be shown to an expert in order for him to be sure about an anomaly's nature. Also ~\cite{sota_human_feedback} includes humans in the AD loop, collecting  experts feedback in an online fashion to improve detection accuracy.

\newcommand{\superquad}[0]{$\qquad\qquad\qquad\qquad\qquad\qquad\qquad\qquad$}
 \begin{figure*}[!t]
 \centering
  \includegraphics[width=0.325\textwidth]{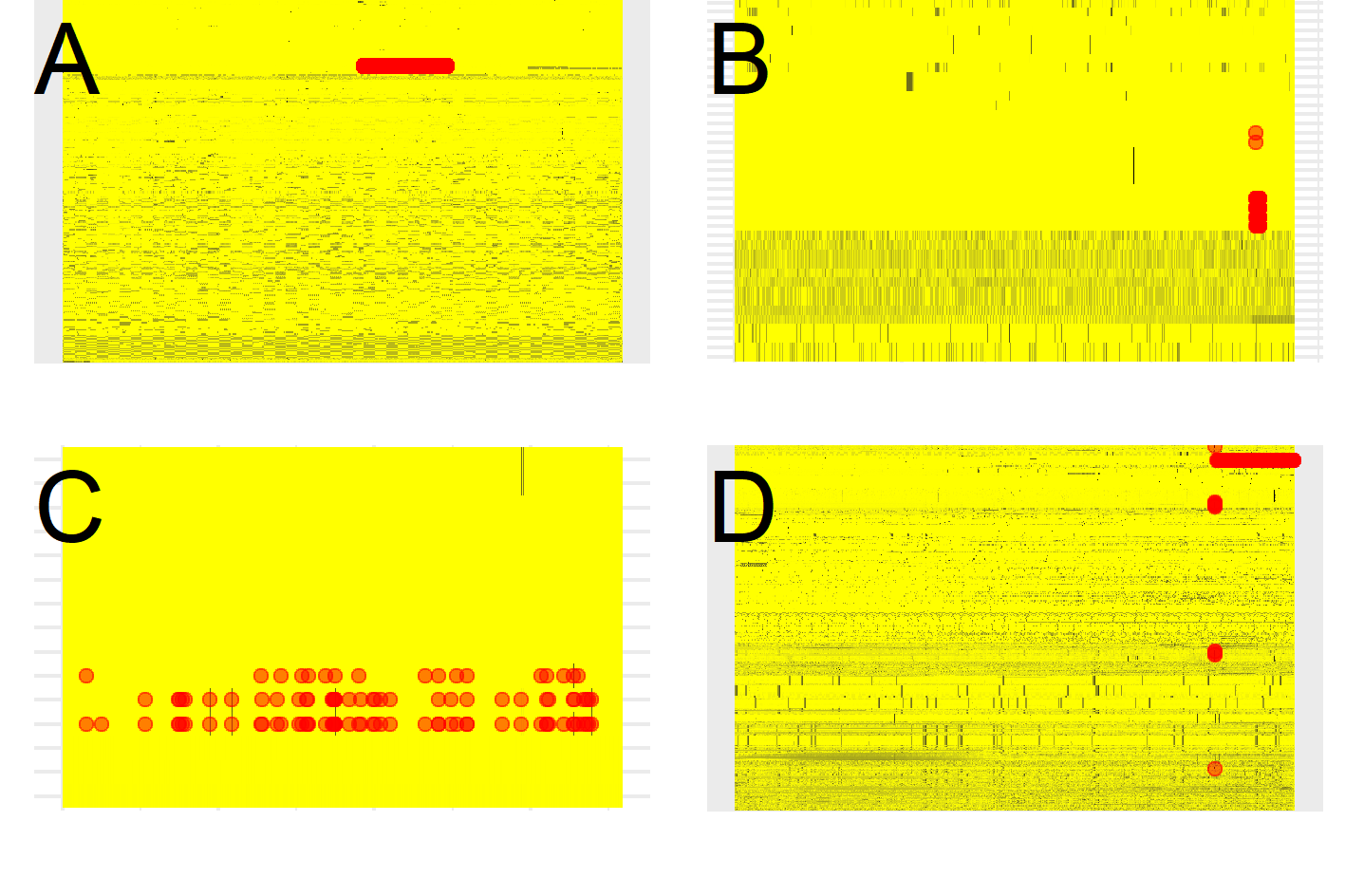}
  \includegraphics[width=0.325\textwidth]{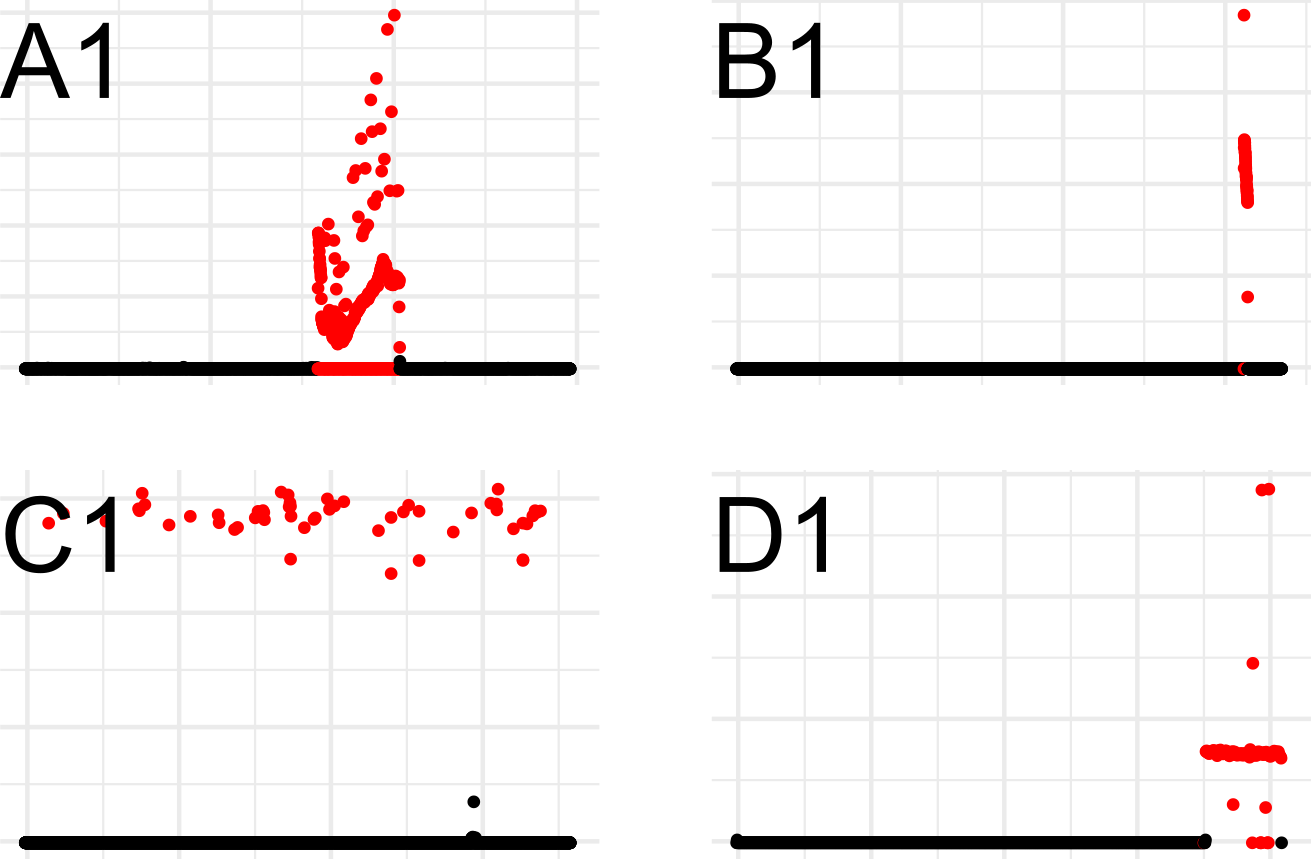}
  \includegraphics[width=0.325\textwidth]{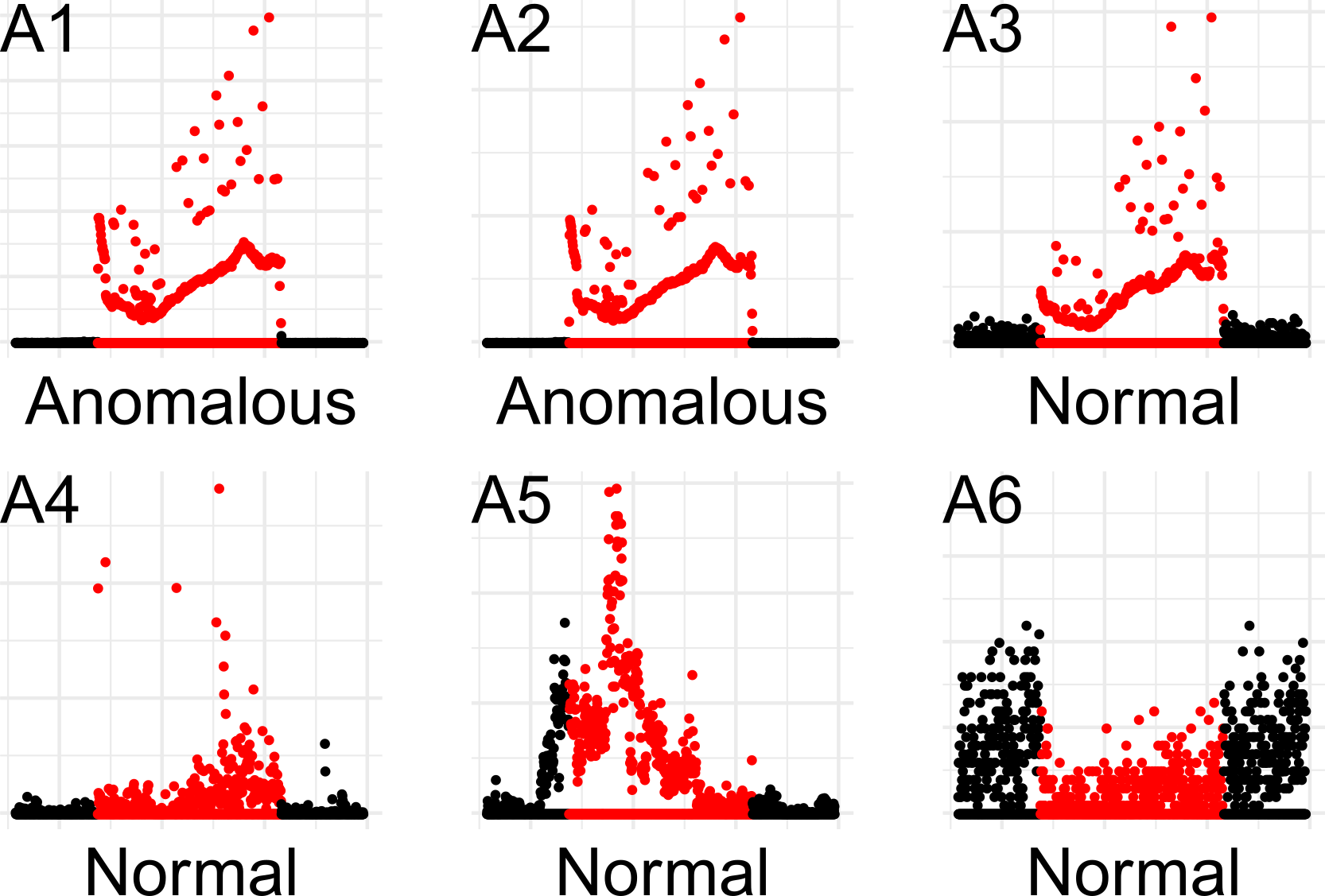} 
(a)\superquad(b)\superquad(c)
  \caption{Compact view of anomalies and ground-truth in our datasets: (a) heatmap of KPI over time, with annotated anomalous ground truth (red dots) for four example datasets A-D;  (b) time evolution of the topmost anomalous feature for the example datasets A-D (anomalous period in red); (c) zoom of the topmost 6 anomalous features of dataset A, of which only the first 3 are flagged by the expert.}
  \label{fig:datasets}
\end{figure*}

While similar in spirit, several differences arise in the interaction mode (explicit~\cite{sota_human_opprentice,sota_human_sfe,sota_human_feedback} vs absent in HURRA), methodology (iterative~\cite{sota_human_opprentice,sota_human_sfe,sota_human_feedback} vs one-shot; supervised~\cite{sota_human_opprentice} vs unsupervised) and final output (sorted timeslots~\cite{sota_human_feedback} vs sorted features). 
Particularly, by seamlessly exploiting ticket solution while avoiding any explicit additional interaction, HURRA improves accuracy and reduces deployment barriers and training costs (as human interface is unmodified).

\section{Dataset Description}\label{sec:dataset}
We validate HURRA leveraging a collection of real datasets, that we describe from Network vs Machine learning angles.

\subsubsection{Network expert viewpoint}
Our datasets represent different router KPIs, extracted from 28 different ISPs via model-driven telemetry, overall comprising almost two months worth of data.  As telcos aim for ``5 nines'' reliability,  the occurrence of anomaly is an unfortunately rare event: hence, these 28 instances represent an ample set of the  troubleshooting tickets occurred during 2019, and each instance has been manually analyzed by a network expert.  The basic properties summarized in Tab.~\ref{table:datasets} testify of a large variability across datasets, of which some example are illustrated  in Fig.~\ref{fig:datasets}. 

Generally, each dataset is collected locally at a single router of a different ISP, so datasets are completely independent from each other. Additionally, there is a large amount of telemetry data (in the order of 70K KPIs)  collected by routers and available for export. At  the same time, exhaustive collection is prohibitive (due to router CPU resources and O\&M bandwidth usage) so that  datasets at our disposal only include the smaller subset of KPIs that have been manually verified by the expert when solving the case (from 6-373 KPIs). 
Overall, datasets comprise 2737 KPIs, of which only 357 are unique: at the same time, datasets are significantly different, since between any pair of datasets there is less than 5\% of common KPIs, and there is not even a single KPI  common  to all datasets. This makes the datasets significantly different from 
work that focuses on data-plane indicators of traffic volume~\cite{munz2008itc} and video-streaming~\cite{fiadino2014itc}, since KPIs in our work are very diverse internal router counters, that pertain to both control and data planes. Also, unlike studies where coarse KPIs are constructed over long timescales (up to 24hours in~\cite{morichetta2018lenta}), the routers in our dataset export KPIs at a fast 1 minute timescale.

The expert labels the time at which anomalies occurred, and additionally writes in his report the KPIs that were found to be responsible of the issue. In this process the network expert can decide \emph{not to report} a KPI as anomalous, irrespectively of the time series behavior, as e.g.,  he  (i) deems other KPIs  more relevant, (ii) believes the KPI to be an effect, and not the root cause of the issue (e.g., in one dataset a large number of packets are dropped when their TTL reaches 0, which is however not flagged by the expert as this symptom is caused by a routing loop, correctly flagged by the expert). The availability of this very fine-grained KPI-level data makes the dataset peculiar: as causality is knowingly hard to assess~\cite{pearl,biersak_causality}, and since our ground truth does not explicitly provide causal information, a natural benchmark for HURRA is thus  the extent of agreement between algorithmic and human judgments.

\begin{table}[!t]
\caption{Summary of basic dataset properties.}
\label{table:datasets}
\centering
\begin{tabular}{ccccc}
\hline
             & \thead{\#Timeslots \\ (rows)} & \thead{\#KPIs \\ (columns)} & \thead{Anomalous \\ Timeslots\%} & \thead{Anomalous \\ KPIs\%} \\ \hline
Minimum      & 211            & 6                 & 0.01\%                 & 0.3\%            \\
1st Quartile & 1025           & 27                & 0.2\%                 & 3.0\%            \\
Median       & 1981           & 73                & 1.2\%                 & 6.0\%            \\
3rd Quartile & 4394           & 126               & 8.2\%                 & 16.7\%           \\
Maximum      & 10770          & 373               & 64.0\%                   & 37.5\%  \\ \cline{1-5}
Total        & 85787          & 2737              & 7.5\%                 & 5.2\%                                     
\end{tabular}
\end{table}

\subsubsection{Machine learning expert viewpoint}
Both the collected features (KPIs), as well as the binary ground truth labels can be abstracted as multivariate time series.  On the one hand, since datasets are independently collected and the set of collected features is also largely varying,  supervised techniques such as those exploited in \cite{casas2016machine} are 
not applicable.  On the other hand, the limited number of features reduces the problems tied to the ``curse of dimensionality''. Overall, these observations suggest  unsupervised algorithms to be a good fit.

For the sake of clarity, we  compactly visualize the diversity of our datasets in
Fig.~\ref{fig:datasets}-(a), which depicts 4 samples datasets A--D  as a heatmap (x-axis represents the time, the y-axis a specific feature, and the z-axis heatmap  encodes the standardized feature value, with darker yellow colors for greater deviation from the mean value), with annotated ground truth (red dots). The figure additionally depicts the temporal evolution of  (b) the topmost anomalous feature found by HURRA in each of the 4 samples datasets (normal portion in black,  anomalous portion in red) as well as of (c) the topmost anomalous 6 features for dataset A. Fig.~\ref{fig:datasets}-(a) and (b) clearly illustrate the dataset variability, even from the small sample reported, in terms of the type of outliers present in the data. 

Fig.~\ref{fig:datasets}-(c) further illustrates the semantic of the available ground truth: it can be seen that the top-three features (A1, A2 and A3) are highly correlated and are also widely varying during the anomaly period; yet the subsequent features (A4, A5 and A6) also present signs of temporally altered behavior, yet these time series were investigated by the human operator, but ruled out in the ticket closing the solution -- well highlighting the cause/effect mismatch.

\section{Methodology}\label{sec:methodology}
The main objective of HURRA  is to  sort the KPIs in a multivariate time series in a way  that reduces the time it takes to complete its troubleshooting process. We first introduce some necessary notation  (Sec.\ref{sec:definition}), and detail the individual building blocks  (Sec.\ref{sec:ad} through \ref{sec:update}).

\subsection{Definitions}\label{sec:definition}
With reference to the building blocks earlier introduced at high-level  in  Fig.~\ref{fig:birdeye}, we denote the collected multi-dimensional time-series data in  as a matrix  $x\in \mathbb{R}^{F T}$, with $F$ the number of existing features and  $T$ the number of timeslots in the dataset. After \circled{0} preprocessing   $\hat x = f(x)$   the data becomes an input  to the \circled{1} \emph{Anomaly Detection (AD)} function,  whose output is a binary vector $a = AD(x) \in \{0,1\}^T$ indicating if the $i$-th timeslot is anomalous or not. The \circled{2}  \emph{Feature Scoring (FS)} function takes both $\hat x$ and $a$ as input, producing a real-valued vector $s = FS(\hat x, a)  \in \mathbb{R}^F$ where $s_j$ represents  an ``anomalous score'' associated to the  $j$-th feature.
An \circled{3} \emph{Expert Knowledge (EK)} function can be used to alter the feature scoring vector $\hat s = EK(s) \in \mathbb{R}^F$ whenever available, else $\hat s = s$.
In this paper, we use \circled{4} \emph{Ground Truth (GT)} labels to assert the performance of our system, as well as to \circled{5} mimic the update of the expert knowledge (more details later). Formally  $g \in \{0,1\}^{F T}$ is a ground-truth matrix, with $g_{jt}=1$  indicating that at the  $t$-th timeslot the $j$-th feature was flagged as anomalous by the expert. It follows that a feature $j$ is  anomalous when $\sum_t g_{jt}>0$, and similarly we can identify anomalous timeslots $a_t$ as the time instants where at least a feature is anomalous, i.e.,  $a_t = \mathbb{1}\left(\sum_j g_{jt}>0\right)$.

\subsection{Data preprocessing and Anomaly Detection (AD)}\label{sec:ad}
Since our focus is on explaining features to a human operator, in this paper we avoid dimensionality reduction techniques that perform linear (e.g., PCA) or non-linear (e.g., t-SNE) data transformation and limitedly perform standard normalization  per-individual KPI $\hat x= (x-\mathbb{E}[x])/\sigma_X$. 

As our our main focus is not to propose yet another anomaly detection algorithm,  we resort to standard and popular unsupervised AD algorithms such as (i) \emph{Isolation Forests} (IF)~\cite{sota_ad_isofor} and (ii) \emph{DBScan}~\cite{dbscan}, as example of partitioning-based and density-based techniques respectively (cfr. Sec.\ref{sec:results} for  hyperparametrization details). We additionally consider an (iii)  \emph{ideal ensemble} method selecting the best among IF and DBScan output,   as well as an (iv) \emph{oracle}   that uses the ground truth labels to identify anomalous timeslots $a_t$. Note that (iii) upper-bounds the performance of a 
\emph{realistic} system combining multiple algorithms, whereas (iv) is ment as the reference performance of an \emph{optimum benchmark}; additionally, the oracle allows to decouple the evaluation of the AD and FS building blocks, i.e., by examining impact of FS under the best AD settings.

\subsection{Feature Scoring (FS)}\label{sec:fs}
The purpose of the FS step is to present human operators with features that are relevant for the solution of the troubleshooting. From a purely data-driven viewpoint, the intuition is that network experts would want to look first at  KPIs  exhibiting drastic change when anomaly happens.

To avoid introducing arbitrary hyperparameters, we aim for non-parametric FS functions. In particular, we define a first $FSa$ function sorting KPIs by their average difference between anomalous and normal times. 
Formally, given $\sum_t a_t$ the number of anomalous timeslots found by the AD algorithm (so that $T-\sum_t a_{t}$ is the number of normal timeslots), the $j$-th feature anomalous score is defined as:
\begin{equation}\label{eq:FSa}
s^{FSa}_{j} = \left| \frac{\sum_t a_{t} \hat x_{jt}}{\sum_t a_{t}}  -  \frac{\sum_t  (1-a_{t})\hat x_{jt}}{T-\sum_t a_{t}}   \right| 
\end{equation}
It is useful to recall that, since $\hat x$ features are normalized, the
magnitude of the scores $s^{FSa}_{j}$ returned by $FSa$  is directly comparable.

We additionally define $FSr$  to measure the difference in the ranking of the $j$-th feature during anomalous vs normal times. Letting normal and anomalous ranks $r^-_j$ and $r^+_j$ respectively:
\begin{eqnarray}
r^+_j &=& \textrm{rank}\left(j : \textstyle\sum_t  a_{t}\hat x_{jt}/\textstyle\sum_t a_{t} \right) \\
r^-_j &=& \textrm{rank}\left(j : \textstyle\sum_t  (1-a_{t})\hat x_{jt}/(T-\textstyle\sum_t a_{t})    \right)
\end{eqnarray}
\noindent the rank-based score becomes:
\begin{equation}\label{eq:FSr}
s^{FSr}_j =  \left| r^+_j -  r^-_j \right|
\end{equation}
Intuitively, whereas $FSa$ compares variations of normal vs anomalous features values in the normalized \emph{feature value} domain, $FSr$ compares the relative impact of such  changes in the \emph{features order} domain -- somewhat analogous to Pearson vs Spearman correlation coefficients.

\subsection{Expert Knowledge (EK)}\label{sec:ek}

By leveraging previously solved cases, one can easily build an EK-base.
Intuitively, if for cases where KPIs $A$, $B$ and $C$ have similar behavior (and thus similar FS scores), only KPI $B$ is labeled as anomalous, 
we can 
assume that KPI $B$ is the only one that experts are interested in seeing to correctly diagnose the case: a ranking more useful for the human operator could be obtained by altering the results of the FS block, e.g. reducing $s_A$ and/or increasing of $s_B$. 
Tracking for each KPI\footnote{Clearly a given feature named $J$ does  not maintain  the same index $j$ across different datasets; however, for the sake of simplicity we prefer to abuse the  notation and  confuse the feature name and index name in this sub-section} $j$ the number $n_j$ of troubleshooting cases where $j$ was observed, as well as the number of times $n^+_j$ it was flagged as anomalous, one can gather its anomaly rate:
\begin{equation}
K^+_j=n^+_j/n_j    \label{eq:splus} 
\end{equation}
Additionally,  EK  can also track, over all datasets, the  number $n^-_j$ of cases where feature named $j$ is \emph{not} flagged by the expert ($g_{jt}=0, \forall t$), despite its score during the anomalous period being larger than the minimum score among the set $\mathcal{A}$ of other features flagged as anomalous:
\begin{equation}
s_j > \min\limits_{k\in \mathcal{A}} s_k,  \quad  \mathcal{A} = \left\{k : \textstyle\sum_t g_{kt}>0 \right\}  \label{eq:sminus}
\end{equation}
Explicitly considering the fact that the expert actually ``ignored'' feature $j$ at a rate  $K^-_j=n^-_j/n_j$ can be helpful in reducing the attention on features that are considered to be less important by the expert (e.g., as they may be effects rather than causes).
The knowledge base can be queried to alter the scores of the FS step as in:
\begin{equation}
\hat s_j= s_j (1+\gamma^+ K^+_j -\gamma^- K^-_j)
\label{eq:fs}
\end{equation} 
by positively biasing ($K^+_j$) scores of KPIs that were found by experts to be the culprit in previous cases, and negatively biasing ($K^-_j$) those that were not flagged by the expert despite having a large anomalous score. In (\ref{eq:fs}) the free parameters $\gamma^+,\gamma^-\in\mathbb{R}^+$  allow to give more importance to past decisions ($\gamma >  1$) or to the natural scoring emerging solely from the current data ($\gamma \rightarrow 0$). 

Interestingly, the frequentist approach can assist in the resolution of common problems (where the $K^+_j$ rate is high and the number of observation $n_j$ is equally large), whereas in uncommon situations (where a new KPI is the culprit), it would still be possible to ignore EK suggestions by reducing $\gamma \rightarrow 0$ (requiring a slight yet intuitive addition to the human interface design). De facto, the use of EK alters FS scores in a semi-supervised manner (and so is able to improve on known KPIs), whereas it leaves the AD block unsupervised (and so able to operate on previously unseed KPIs).

\subsection{Update of Expert Knowledge Base}\label{sec:update}
We remark that EK management is particularly lean in HURRA. In case of cold-start (i.e., no previous EK base),  $K^+_j=0$ so that $\hat s_j = s_j$ by definition. Additionally, the system can operate in an incremental manner and is capable of learning over time without explicit human intervention:  the information is extracted from the tickets by simple update rules (\ref{eq:splus})--(\ref{eq:sminus}).
Finally, the EK update mechansim supports tranfer learning, as it is trivial to ``merge'' knowledge bases coming from multiple ISPs, by e.g., simple weighted average of $K^+_j$ and $K^-_j$ rates. As a matter of fact, our datasets already aggregates multiple ISPs deployment, which can therefore prove (or invalidate) the transferability of our proposed semi-supervised mechanism.

In this work, we simulate the EK update process by performing \emph{leave-one-out} cross-validation: i.e., upon analysis of a given dataset, 
we only consider knowledge coming from the other datases, which prevents overfit. 
Also, in reason of the significant heterogeneity (very few KPIs in common across datasets), and the relative low number of datasets, the evaluation in this paper constitute a  stress-test w.r.t. the expected performance of a massively deployed EK system having collected thousands of instances.

\begin{figure}[t]
\begin{center}
  \includegraphics[width=0.75\columnwidth]{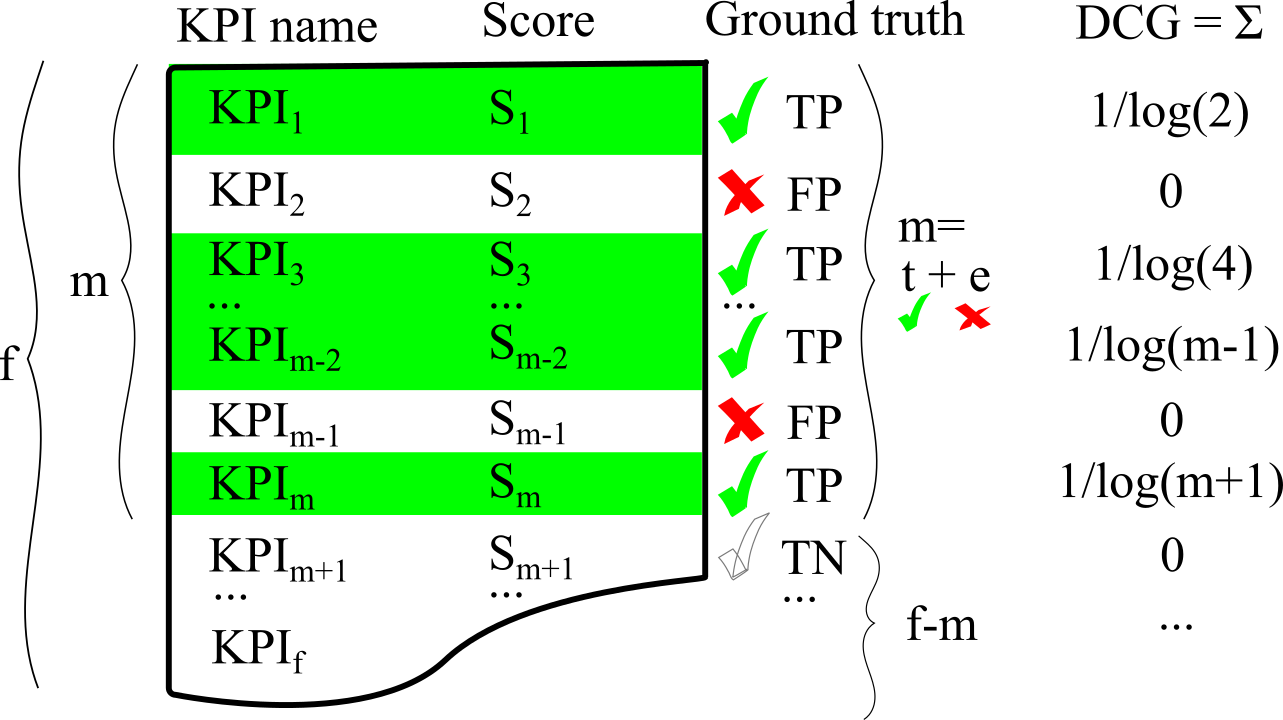}
  \caption{Synoptic of the performance metrics for attention focus mechanism.}
  \label{fig:synoptic}
\end{center}
\end{figure}

\section{Experimental Results}\label{sec:results}
We now define the metrics (Sec.\ref{sec:res:metrics}), that we use to illustrate the performance at a glance (Sec.\ref{sec:res:glance}) before delving into the contribution of each building block (Sec.\ref{sec:res:fs} through \ref{sec:res:ek}) and putting them in perspective  (Sec.\ref{sec:res:summary}).

\subsection{Performance metrics}\label{sec:res:metrics}
We define metrics to assess the extent of agreement among the ranking induced by $\hat s$ and the ground truth $g$, defined by the expert, that has observed and judged $f$ features to find the $t<f$ anomalous ones, with the goal of minimizing the number $e$ of irrelevant features that are presented to the expert.

\subsubsection{Network expert viewpoint}
Fig.~\ref{fig:synoptic}  illustrates an example output where the $f$ features are sorted according to their decreasing anomalous score $s$ so that $s_i>s_j$ for $i<j$.
Ideally, an expert would want an algorithm to  present him these $t$ most relevant features at the top of the ranking. 
In particular, we  denote with $m = t + e$ the last position occupied in our ranking by a feature that the expert has flagged as relevant to troubleshoot the issue. By definition, all features occupying a rank higher than $m$ are non-anomalous (true negatives), whereas $e$ KPIs within the $m$ topmost returned by the ranking are false positives (feature 2 and $m-1$ in the example of Fig.~\ref{fig:synoptic}). Formally we define the \emph{Reading effort}, $m = t+e$ as the number of KPIs $m$ that must be examined by an expert using HURRA to view all $t\le m$ anomalous KPI.

 \begin{figure*}[t]
 \centering
  \includegraphics[width=0.4\textwidth]{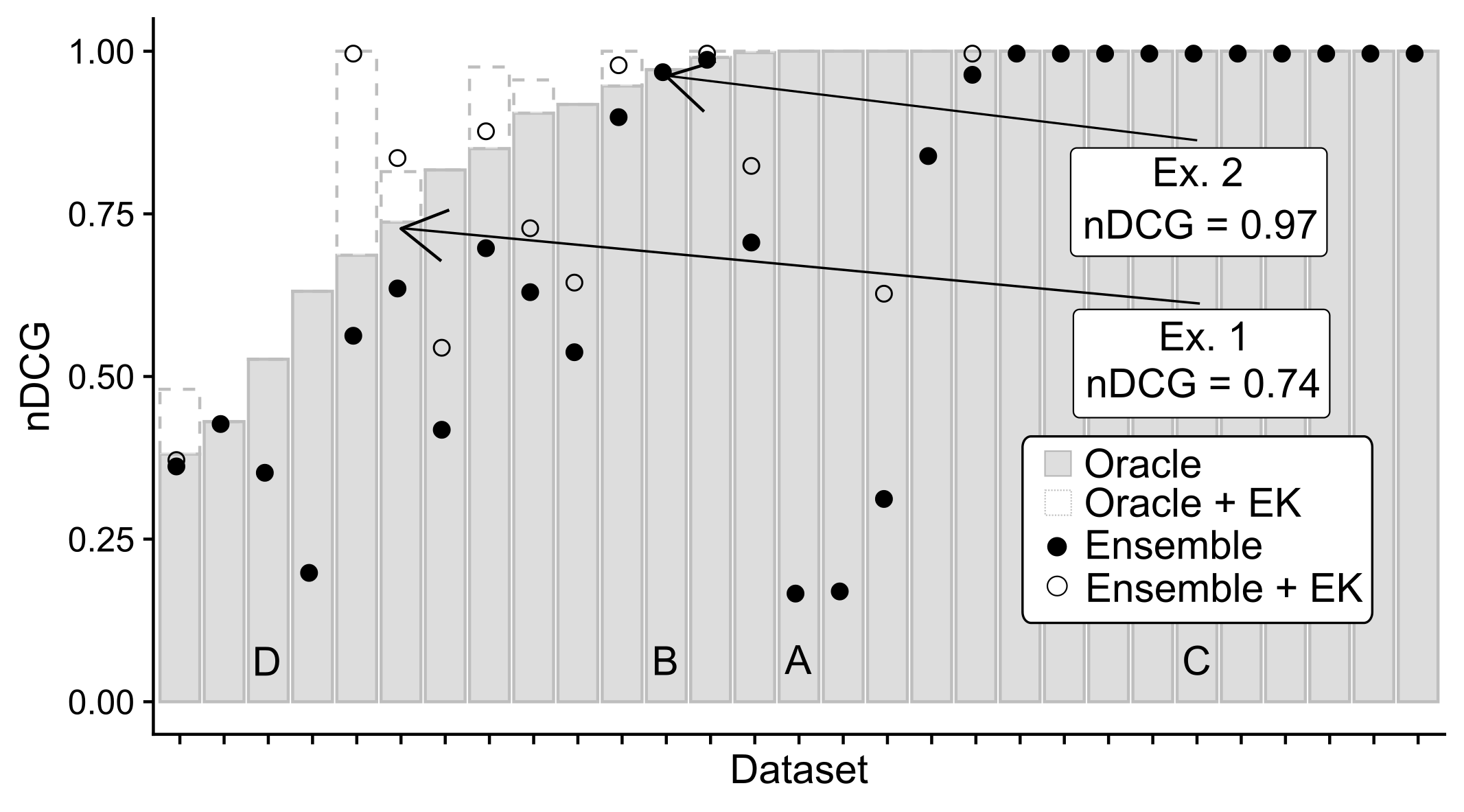}
  \raisebox{0.8cm}{\fbox{\includegraphics[width=0.28\textwidth]{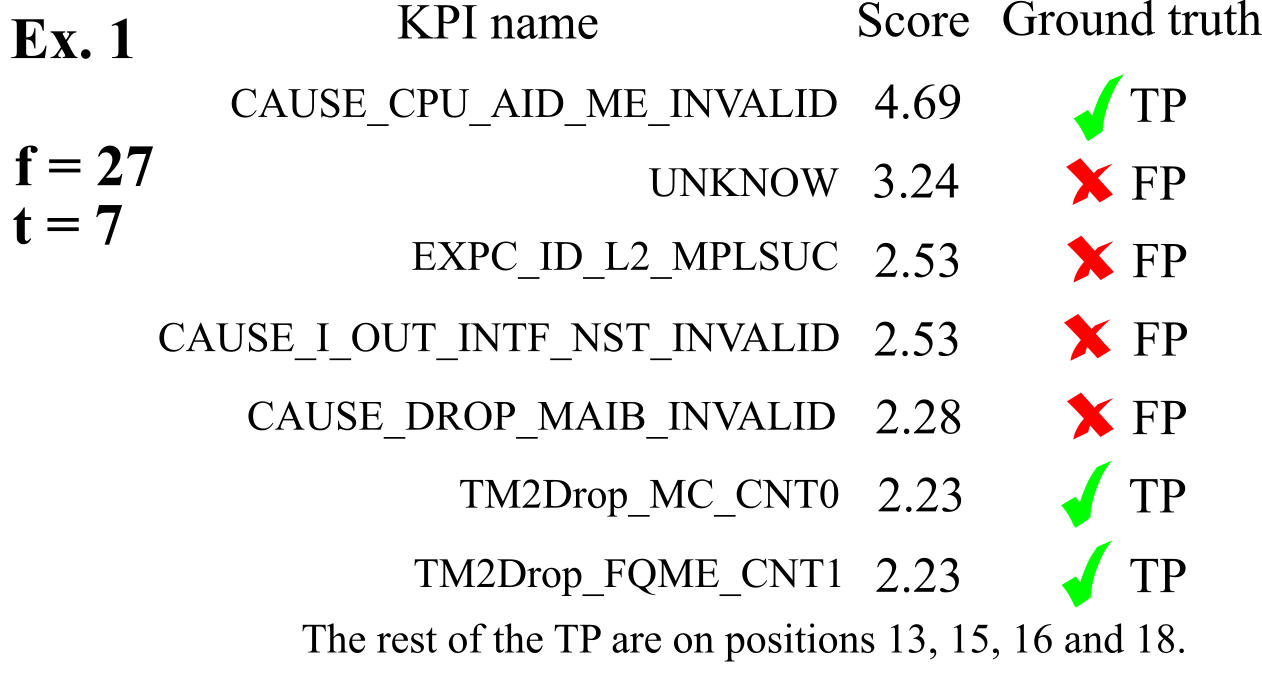}}}
  \raisebox{0.8cm}{\fbox{\includegraphics[width=0.28\textwidth]{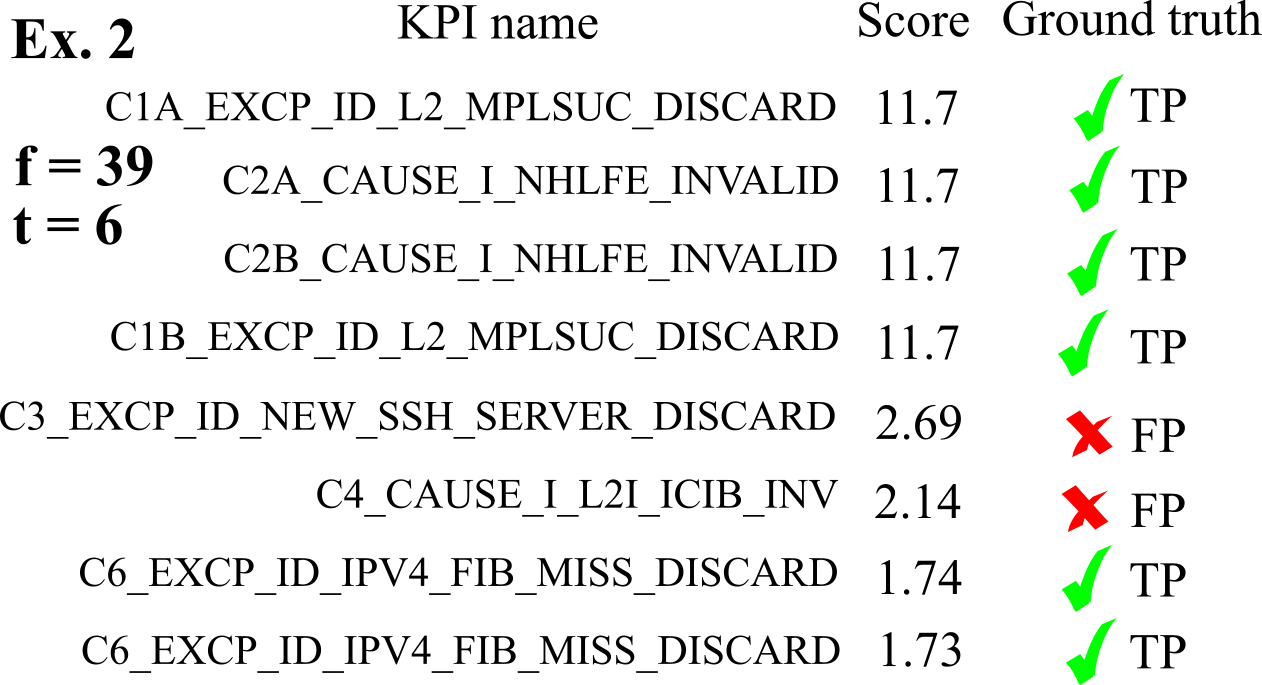}}}
  \caption{Experimental results at a glance: nDCG  for oracle (grey bar) vs ensemble (black points) across all datasets (left) and  example sorting for two nDCG values (right).}
  \label{fig:glance}
\end{figure*}

\subsubsection{Machine learning viewpoint}
From a machine learning viewpoint, a robust metric to measure the extent of agreement between the expert ground truth $g$ and the output ranking induced by $s$ is represented by the normalized Discounted Cumulative Gain (nDCG)  which is commonly used in information retrieval to compare rankings~\cite{yahoo_rankings_from_xgboost} and is defined as the ratio of two factors:
\begin{equation}
nDCG=DCG/iDCG
\end{equation}
The ideal DCG score (iDCG) is attained when all KPIs flagged as anomalous by the expert appear in the first $t$ positions (the order of non-anomalous KPIs after $t$ is irrelevant):
\begin{equation}
iDCG = \sum_{i=1}^{t<n}\frac{1}{log_2{(i+1)}}
\end{equation}
\noindent The DCG score of any ranking can be computed by accounting only for the anomalous KPIs:
\begin{equation}
DCG = \sum_{i=1}^{n}\frac{\mathbb{1}( \sum_t g_{it} > 0 )} {log_2{(i+1)}}
\end{equation}
\noindent where it should be noted that the KPI returned in the $i$-th position by a FS policy may not be present in the ground truth  (false positive in Fig.\ref{fig:synoptic}) in which case it has null contribution to DCG, as well as it lowers contribution of the next true positive feature. nDCG is  bounded in [0,1] and equals 1 for perfect matching with the expert solution: although less intuitive, this metric allow to unbiased comparison across datasets, as well as quickly grasping the optimality gap of the HURRA solution.

\subsection{Performance at a glance}\label{sec:res:glance}
We start by illustrating  performance at a glance over all datasets in Fig.\ref{fig:glance}.  The picture reports the nDCG scores obtained  using the best FS function across all datasets (ranked by increasing nDCG), when the AD task is performed by an Oracle (grey bar, upper bound of the performance) or by an  Ensemble of unsupervised methods (black points, realistic results)
with (light points and no shading) or without (dark points and shading) Expert knowledge.
It can be seen that, in numerous troubleshooting cases, the ensemble is able to attain high levels of agreement with the expert (the dots approach the envelope). It also appears that leveraging expert knowledge can assist both the ensemble as well as the Oracle, which is interesting. Finally, it can also be seen that there are few cases that are particularly difficult to solve (nDCG$<$0.25 in 3 cases, where the number of anomalous/total features is 1/157, 2/335 and 1/346 respectively): anomaly detection appears difficult for the algorithms we considered due to the curse of dimensionality, so subspace-based methods~\cite{mazel2011sub,sota_hics,sota_lookout} would be more appropriate in these cases (cfr. Sec.\ref{sec:discussion}).

To assist the reader with the interpretation of the nDCG metric, Fig.\ref{fig:glance} additionally reports the ranking produced by HURRA in two examples datasets, that portray cases of moderate and high nDCG scores. 
For instance, in the high-nDCG case (Ex. 2), it can be seen that the $t=6$ anomalous out of $f=39$ total features are within a reading effort of $m=8$ KPIs with $e=2$ false positive for an overall nDCG$=0.97$. Similar observations can be made for the low-nDCG case (Ex. 1), where there is a higher number of false positive ($e=11$).

\begin{figure}[t]
 \centering
  \includegraphics[width=0.5\textwidth]{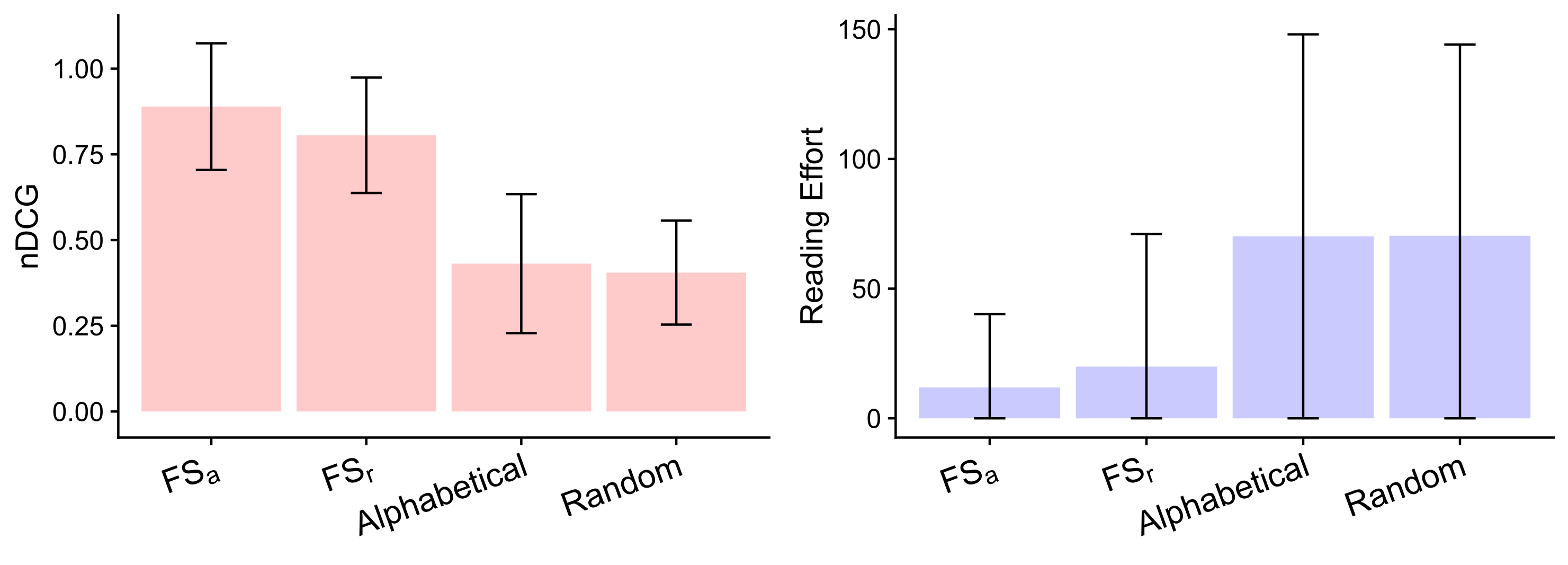}
  \caption{Impact of Feature Scoring on nDCG and reading effort metrics (average and standard deviation).}
  \label{fig:FS}
\end{figure}

\subsection{Impact of Feature Scoring (FS)}\label{sec:res:fs}
Fig.\ref{fig:FS} next contrasts FS performance for nDCG and reading effort metrics considering four different ranking policies.
Two rankings are determined by the (i) average-based $FSa$  and  (ii) rank-based $FSr$ feature scoring defined by (\ref{eq:FSa}) and (\ref{eq:FSr}) respectively.
We further add two na\"ive rankings that are  evaluated as the theoretical expectation of an (iii) \emph{alphabetical} feature ranking, that represents the current system  and (iv) a \emph{random} ranking, for the sake of comparison.

It can be seen that $FSa$ and $FSr$ reduce the reading effort of over 50 features on average, and increase nDCG by roughly a factor of two with respect to  alphabetical (or random) sorting.  The fact that alphabetical and random sorting have close performance can be also interpreted as an indication that presenting KPIs in alphabetical order has no value for the troubleshooting expert.

Finally, notice that $FSa$  is slightly more accurate then $FSr$: since the $FSr$ is also computationally more costly (as it requires, in addition to the final sorting, to perform two additional sorting to determine rankings $r^+$ and $r^-$ that are needed for computing the score), we limitedly consider $FSa$ in what follows.

\subsection{Impact of Anomaly Detection (AD)}\label{sec:res:ad}
\begin{figure}[t]
  \centering
  \includegraphics[width=0.48\textwidth]{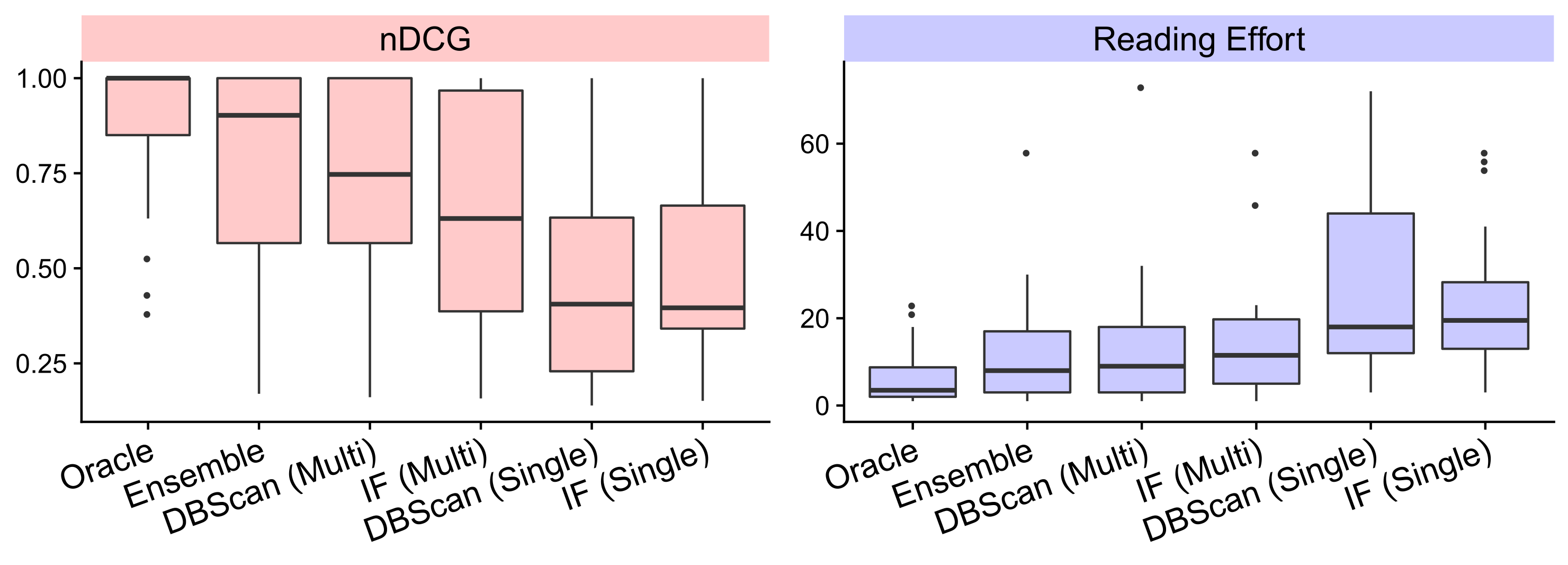}
  \caption{Impact of Anomaly Detection on nDCG and reading effort.}
  \label{fig:AD}
\end{figure}

Fig.\ref{fig:AD} further allows to grasp the impact of AD choices, using average-based feature scoring $FSa$. Particularly, boxplots report the quartile of nDCG and reading effort for, from left to right (i) the anomaly detection oracle, (ii) an ideal ensemble combining best IF and DBScan results
(iii)-(iv) the best IF and DBScan results using a \emph{specific hyperparametrization} for each dataset and (v)-(vi) IF and DBScan results using a \emph{single hyperparametrization} for all datasets. 
In particular, for DBScan the best hyperparametrization for point (iii) is found by grid optimization of $(\epsilon,minPts)$ settings  with  $\epsilon \in \{1,\ldots,20\}$
and $minPts \in \{2,5,10,20,40,\ldots,200\}$, for a total of 260 combinations. Out of these hyperparametrizations, the setting $(\epsilon,minPts)=(13,80)$ yields to  the best average nDCG over all dataset, and is used for (v).
For IF, the number of trees is fixed to 300 and we explore three options, for a total of 9 hyperparametrizations:  (a) Static contamination factor set to the top X\% of the samples, with $X\in$: $\{0.1\%, 1\%, 5\%, 10\%\}$; (b) Static threshold $\theta_S$ on isolation score $\theta_S \in \{0.55, 0.6, 0.65, 0.7\}$; (c) Dynamic threshold $\theta_D$, set by finding an elbow in the topmost 10\% of isolation scores. Any of the above can be used for (iv) whereas a static contamination factor set to $1\%$ yields the best average nDCG over all dataset and is used for (vi).

It is useful to observe how performance degrades moving away from the oracle (i) from left to right. Contrasting (i)-(ii) it can be seen that, while the nDCG is reduced, the reading effort remains moderate; this hints to the fact that while the number of false positive $e$ KPIs is low, these may occupy positions towards the top of the ranking, which have a more pronounced effect on nDCG.
Further considering (iii) multiple hyperparametrization of DBScan, it can be seen that while the median nDCG drops, its performance is still satisfactory.
Similar considerations can be extended, to a lesser extent, to (iv) Isolation Forest, yet the main difficulty for (iii)-(iv) lies in setting hyperparameters in a completely unsupervised manner.
Conversely, a single hyperparametrization (v)-(vi) is clearly insufficient, 
as it constitutes a too small improvement with respect to the current 
na\"ive alphabetical ranking (recall Fig.\ref{fig:FS}) to be even worth considering for practical deployments.

We further expand, for each dataset, the different nDCG performance that can be attained depending on the selected hyperparametrization. Results for all  hyperparametrization explored are reported in  Fig.\ref{fig:AD},  highlighting in red the unique combination yielding the best  average performance across all datasets.
It can clearly be seen that, for any given dataset, performance can vary widely according to the hyperparametrization.
As the gap between single and multi-hyperparametrization shows, there is also little hope that classic tuning approaches 
(e.g., use the system with a single hyperparametrization coming from a state-of-the art cross-fold validation result of this study) would work in practice. As such, automatically finding an hyperparametrization that produces good results for the case under observation has high practical relevance: we discuss preliminary comforting results on this critical point in Sec.\ref{sec:discussion}.

 \begin{figure}[t]
 \centering
  \includegraphics[width=0.5\textwidth]{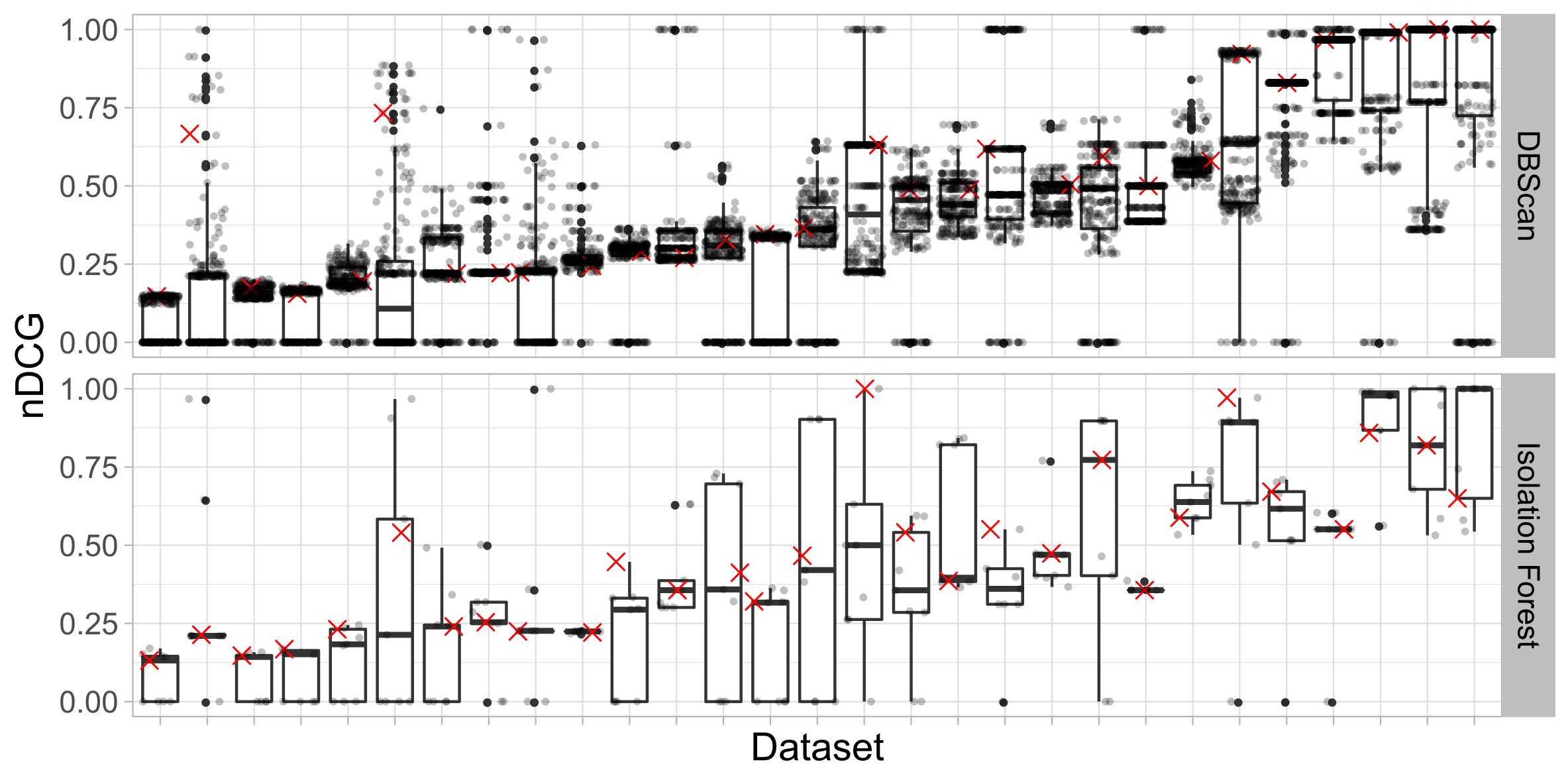}
  \caption{Impact of AD Hyperparametrization  on nDCG: DBScan (top) and IF (bottom). The red cross indicates the performance of the fixed setting performing best on average over all datasets.}
  \label{fig:hyperparam}
\end{figure}

\subsection{Impact of Expert Knowledge (EK)}\label{sec:res:ek}
\begin{figure}[t]
\centering
  \includegraphics[width=0.4\textwidth]{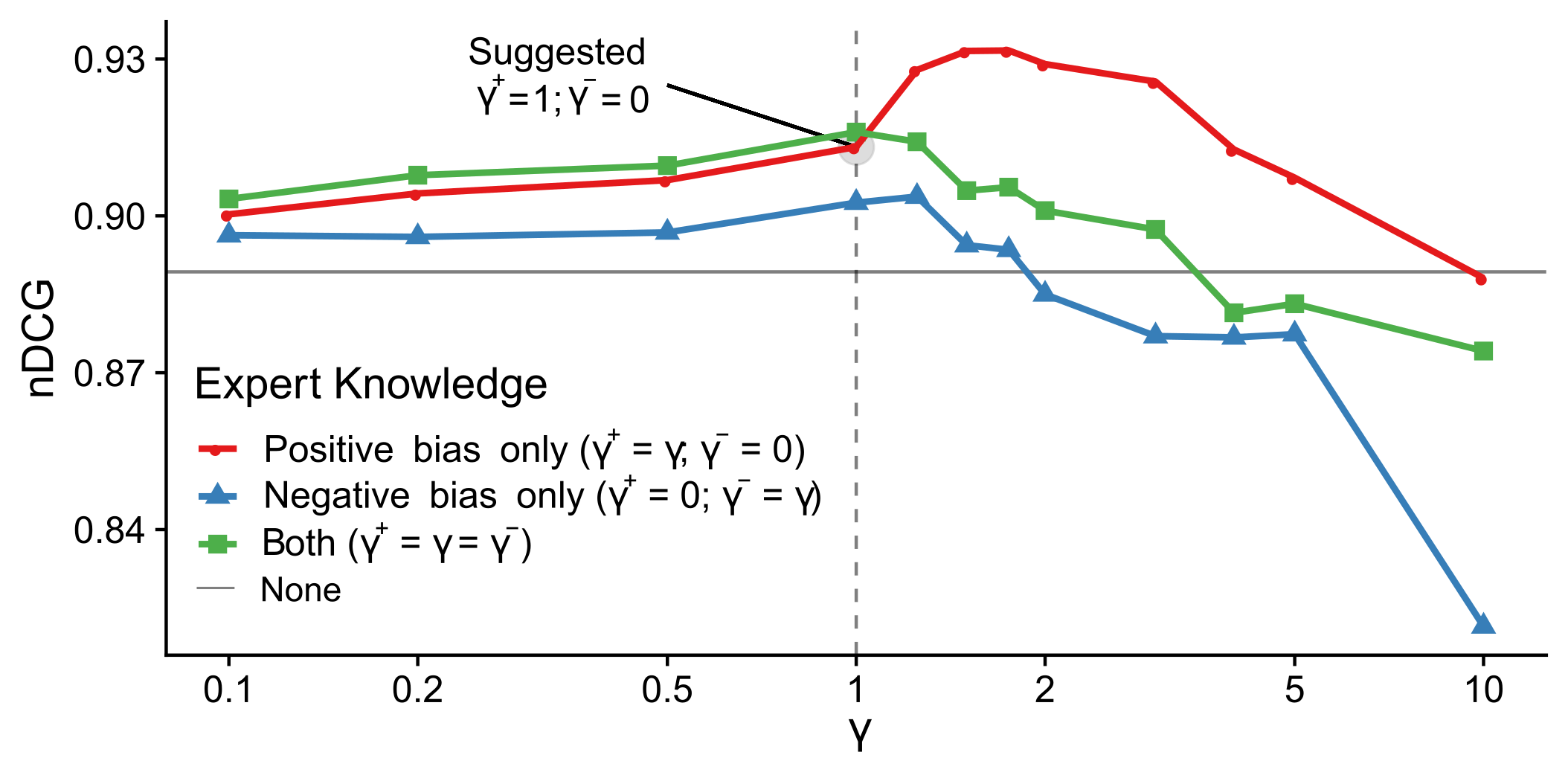}
    \caption{Impact of Expert Knowledge of nDCG on the anomaly detection oracle.}
  \label{fig:EK}
\end{figure}

We next gauge the value of the EK block in Fig.\ref{fig:EK}, which analyzes the impact of leveraging expert knowledge on top of the anomaly detection Oracle and average-based $FSa$ (the latter, represented as a solid horizontal black line in the picture corresponding to an average nDCG$=0.89$) over all datasets, 
as  a function of the EK gain $\gamma$. Clearly, improving such a high nDCG should prove quite difficult, and this should thus be considered as a conservative scenario to assess benefits deriving from EK (which is already made difficult by the heterogeneity of the datasets).

We consider several cases to better isolate the effects of positive and negative bias, particularly (i) positive only $\gamma^+=\gamma$ and $\gamma^- = 0$, (ii) negative only 
 $\gamma^+=0$ and $\gamma^- =\gamma$ and (iii) both effects with equal gain $\gamma^+= \gamma^- =\gamma$.
Several interesting remarks are in order. First, observe that improvements in the ranking are already evident for moderate gain $\gamma \ge 1/10$. Second, observe that positive bias $K^+_j$ has a stronger effect than negative one $K^-_j$, yet the effects bring a further slight benefit when combined. Third, the benefit of positive bias tops around $\gamma=2$ (in these datasets) and remains beneficial even for very large gains $\gamma\approx 10$.
Fourth, the effects of negative bias can instead worsen the resulting rank for $\gamma>2$ so that the combination of positive and negative bias (with equal weight) is also affected for large gains.

These different effects are easy to interpret considering the different nature of positive and negative biases. In particular, positive bias $K^+_j$  assesses the rate at which observers individually flag KPI $j$, as observed by independent observers over multiple independent datasets. Whereas $K^+_j$ does not directly translate into the probability that KPI $j$ is also anomalous in the dataset under observation, increasing the score proportionally to it  can help shifting the attention  toward KPIs that are often considered valuable toward the  troubleshooting solution.
Interestingly, as the the support of the observation grows, the metric becomes more accurate, and as the bias grows, more frequent problems can be solved more effectively. 

In contrast,  the computation of negative bias $K^-_j$ couples observations across several metrics, as the numerator in the  expression relates to the number of observations where the KPI $j$ is flagged as normal despite its score $s_j$ exceeds the score of at least one KPI flagged as anomalous. However, the subset of KPIs in common among any pair of dataset is small, so the knowledge distilled in $K^-_j$ appears more difficult to transfer across ISPs.

In summary, (i)  it appears that simple frequentist representation of EK are providing measurable advantages, even on top of oracle anomaly detection with high nDCG; (ii) gain from positive bias $K^+_j$ is more robust than $K^-_j$ and is furthermore consistent for very large parameter range $\gamma\in[0,10]$;
(iii) overally,  we  recommend the case of EK gathered with a positive bias only $\gamma^-=0$ and a limited gain $\gamma^+=1$, which  also conservatively assesses EK benefits.

\subsection{Relative impact}\label{sec:res:summary}
\begin{figure}[t]
  \includegraphics[width=0.48\textwidth]{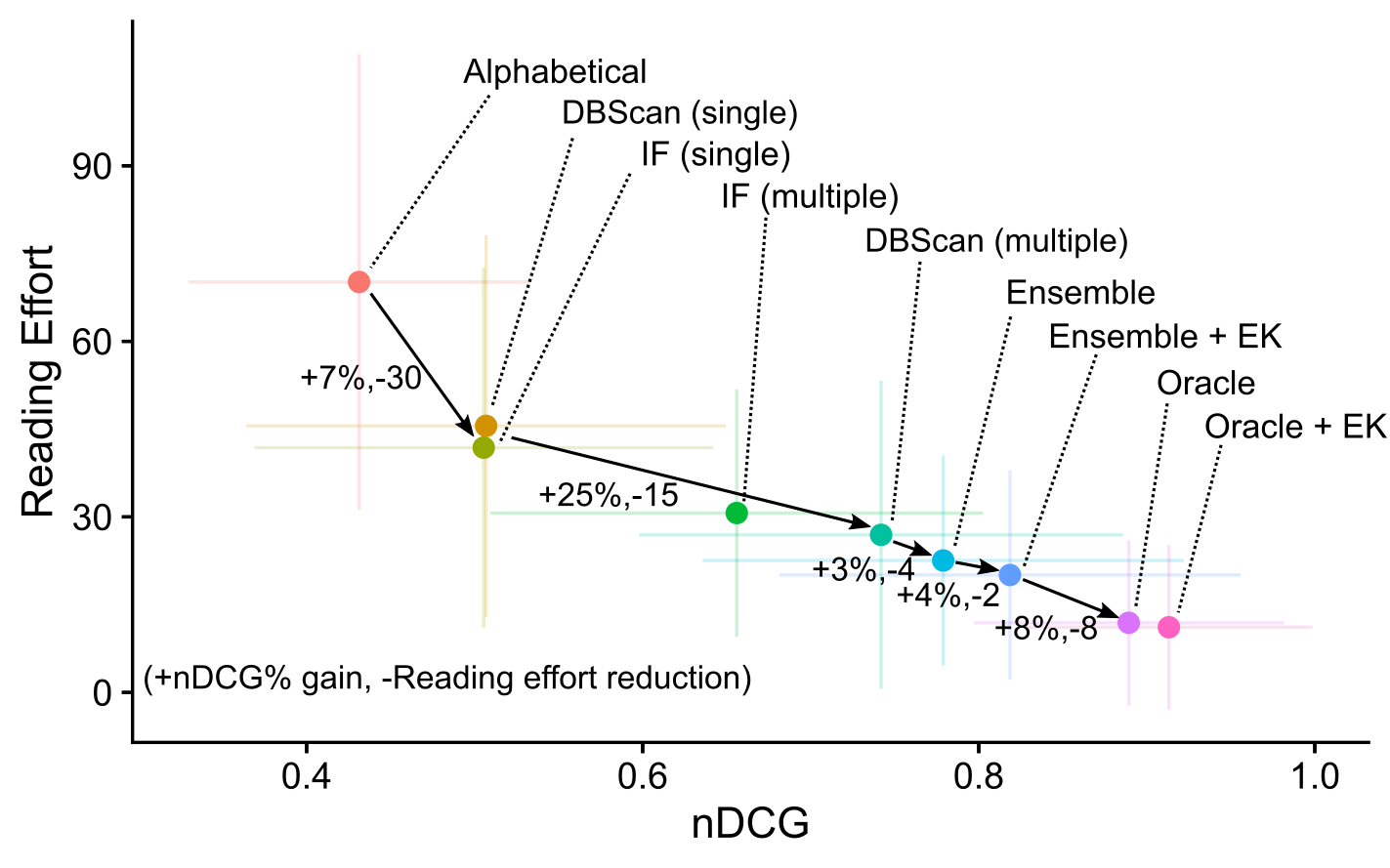}
  \caption{Relative impact of HURRA's building blocks: Scatter plot of nDCG vs Reading effort performance (ideal performance in the bottom right corner).}
  \label{fig:relative}
\end{figure}

Finally, it is insightful to compare the impact of the different building blocks that compose HURRA, which we compactly summarize as a scatter plot in Fig.\ref{fig:relative}.
Performance of the na\"ive  Alphabetical system appears on the top-left corner, far from the bottom-right corner of ideal oracle performance. Improvement in both nDCG (+7\%) and reading effort  reduction  (-30) of the  na\"ive system can already be achieved by deploying unsupervised clustering with a unique hyperparametrization setting (that, however, requires machine-learning experts for a first tuning). The use of multiple hyperparametrizations brings a further significant advantage even considering a single algorithm (+25\% -15 for DBScan and +15\%, -28 for IF, not shown), and of course ameliorate further when results are combined in the ensemble, which is quite close to the oracle.

Finally, the adoption of expert knowledge, even in the very simple form presented in HURRA, is beneficial not only in  case of ensembles (+4\%, -2) but also when an Oracle solves the AD problem (+3\%, -1) although with diminishing returns (since the oracle performance are already close to perfect agreement with the expert, recall Fig.\ref{fig:EK}). The EK gain even in the presence of an oracle for anomaly detection can be explained with the fact that not all KPIs with a manifest anomalous behavior are important, as they may be a symptom rather than a cause. This  confirms that even a simple frequency-based approach in exploiting expert knowledge can lead the expert in focusing on KPIs that are closer to the ``root cause'', without directly exploiting causality, and that this information can be ``transferable'' in machine learning terms, across datasets.

\section{Discussion}\label{sec:discussion}

Network troubleshooting is still a human-intensive process:  HURRA not only assists human operators in their tasks (using unsupervised learning techniques), but that can also learn from the final troubleshooting solutions validated by the operator (without requiring his explicit intervention).  HURRA is easily deployable (as it requires minimal changes with respect to UI and no human interaction), hot-swappable (as it does not require an EK base) and is expected to improve over time (as the EK base grows). While performance benefits are clear, the limits of HURRA are also worth pointing out, as they leave interesting open questions.

\fakepar{Stream-mode}  While this work focues on classic batch methods,  a natural next step would be to move to stream-mode operation as in~\cite{dromard2016online}, e.g., by replacing  Isolation Forest with Robust Random Cut Forest, or DBScan with DenStream, which would allow HURRA to not only react on demand, but to also proactively trigger ``human readable'' alerts.

\fakepar{Decoupled AD/FS} The fact that the performance of the FS module is tied to the performance of the AD step is a double-edged sword: on the one hand, if the AD part is solved, the FS is easy to apply and will generate useful insights. However, we have seen few hard-to-solve instances (recall the low nDCG cases in Fig.\ref{fig:glance}) where subspace methods~\cite{mazel2011sub}, coupling FS and AD, would be preferable.

\fakepar{FS Applicability} Given our troubleshooting focus, the proposed FS policies are designed to work well when solving a single ticket. As such, to let the system running continuously, $FSa$ would require to maintain online average of normal samples, whereas $FSr$ would incur prohibitive costs.

\fakepar{AutoML} More generally, our results show that no single AD algorithm with a single hyperparametrization, can be expected to work well for all circumstances. Thus, an unsupervised meta-learning approach recommending the appropriate selection per dataset would be of invaluable importance. 

\begin{figure}[t]
  \includegraphics[width=0.48\textwidth]{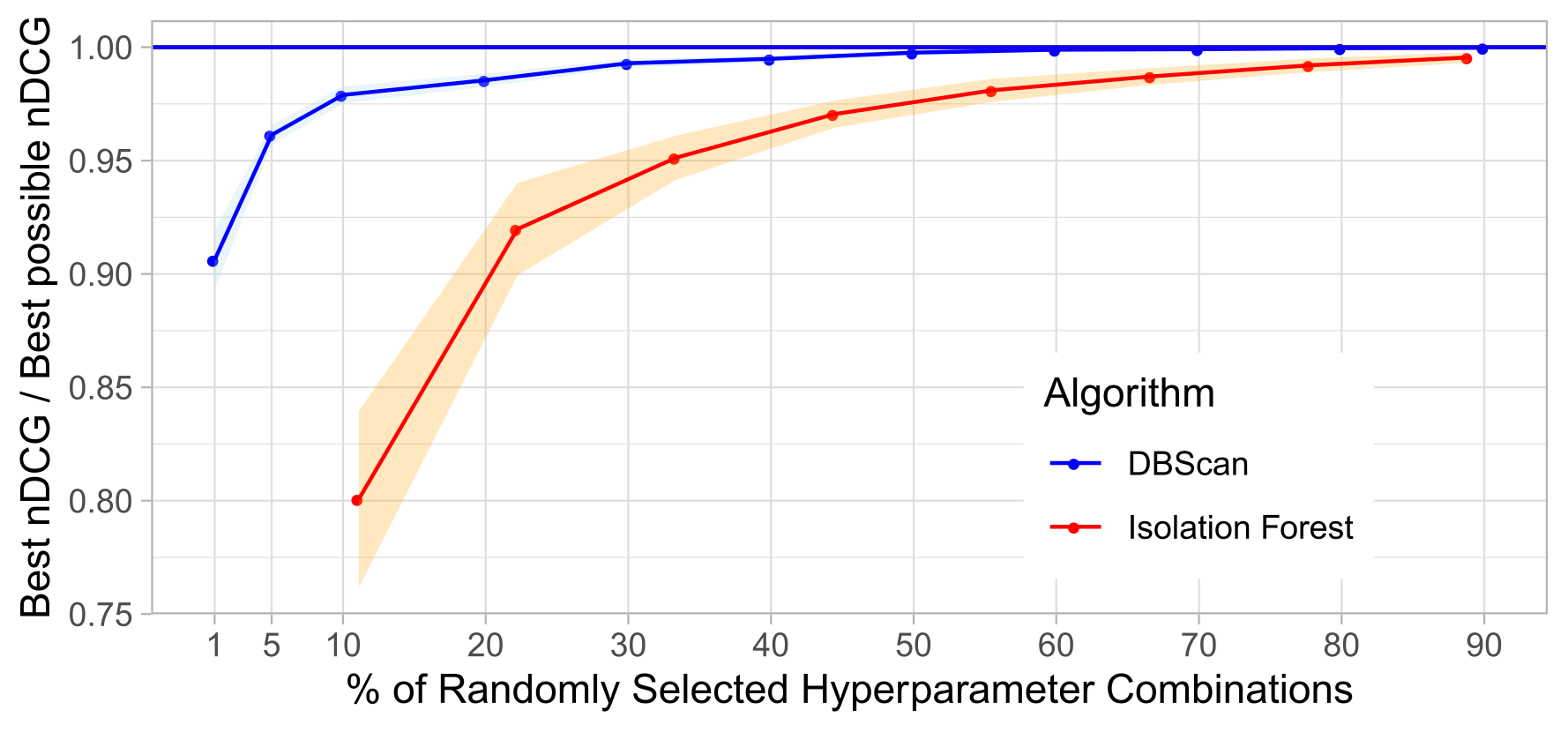}
  \caption{Quality of the solution as a function of the fraction of randomly tested hyperparametrization settings.}
  \label{fig:hyperparam-experiment}
\end{figure}

As the latter point is particularly relevant, we perform a simple experiment to show that random-hyperparametrization would  allow to quickly produce results under (mild) assumption of  being able to  compare, in a binary fashion, which of two solutions is better. In other words, we do not require a precise quantitatively assessment of a solution (e.g., the  value of the gradient toward the optimum), but we assume of being able to roughly assess the quality of the solution (e.g., the \emph{sign} of the gradient suffices). We then automatically tune the algorithm by (i) randomizing an hyperparameter settings and (ii) keeping the most promising between the current settings and the new ones. Fig.\ref{fig:hyperparam-experiment} depicts
as a function of the fraction of the parameter combination explored on the x-axis, the average nDCG (with confidence bars gathered from 100 independent simulations) normalized over the maximum achievable nDCG over the whole explored combinations.
It can be seen that randomized hyperparametrization quickly  converges to over 90\% of the asymptotic nDCG gathered via  exhaustive grid exploration: 3 tests (1\% of the explored combinations) are sufficient in DBScan and 2 tests (20\%) are sufficient in IF, with diminishing returns afterward. Devising an unsupervised randomized strategy able to converge in practice is part of our future work.

\section{Conclusions}\label{sec:conclusions}

This paper presents HURRA, a system designed to reduce the time taken by human operators for network troubleshooting.
 HURRA is simple and modular, with a design that decouples the Anomaly Detection (AD), Feature Scoring (FS) and Expert Knowlege (EK) blocks.   HURRA leverages unsupervised techniques for AD and FS:  our performance evalution shows that whereas single algorithms with fixed hyperparametrization do provide only a very limited benefit, 
 the use of ensembles and multiple hyperparametrization has the potential of making unsupervised systems of practical values.  Furthermore, HURRA is capable of seamlessly building  a  knowledge base EK, by exploiting the information coming from  troubleshooting solutions validated by the human operator, without furthermore requiring his explicit intervention. EK can then be used to refine the unsupervised solution provided by AD+FS blocks.  Numerical results show that such a system is promising, in that the combined use of ensembles and expert knowledge approaches ideal performance achieved by an oracle.


\bibliographystyle{IEEEtran}
\bibliography{refs}

\end{document}